\documentclass{article}
\usepackage{spconf,amsmath,graphicx}
\usepackage{ dsfont }
\usepackage[export]{adjustbox}
\usepackage{hyperref}
\usepackage[all]{hypcap}  

\DeclareGraphicsExtensions{.png}

\title{FA-GAN: FEATURE-AWARE GAN FOR TEXT TO IMAGE SYNTHESIS}
\name{Eunyeong Jeon\qquad Kunhee Kim\qquad Daijin Kim}

\address{Dept. of Computer Science and Engineering, Pohang University of Science and Technology, Korea} 
\begin{document}
%
\maketitle

\begin{abstract}
Text-to-image synthesis aims to generate a photo-realistic image from a given natural language description. Previous works have made significant progress with Generative Adversarial Networks (GANs). Nonetheless, it is still hard to generate intact objects or clear textures (Fig \ref{fig:fig1}). To address this issue, we propose Feature-Aware Generative Adversarial Network (FA-GAN) to synthesize a high-quality image by integrating two techniques: a self-supervised discriminator and a feature-aware loss. First, we design a self-supervised discriminator with an auxiliary decoder so that the discriminator can extract better representation. Secondly, we introduce a feature-aware loss to provide the generator more direct supervision by employing the feature representation from the self-supervised discriminator. Experiments on the MS-COCO dataset show that our proposed method significantly advances the state-of-the-art FID score from 28.92 to 24.58.

\end{abstract}

\section{Introduction}
\label{sec:intro}

Generative Adversarial Networks (GANs) \cite{c1} have led remarkable success in image generation with various types of conditions \cite{c2,c3,c4,c6,c7,c8}. Research on text-to-image synthesis is in the limelight because of the expressiveness of natural language unlike random noises or labels. Nevertheless, it is challenging to synthesize high-quality images while satisfying the diverse constraints of text descriptions. For example, generating \textit{`a few people on skis standing on a mountain top'} is more difficult than generating \textit{`people'}.

Most existing works \cite{c9,c10,c11,c12,c13} have achieved remarkable progress by proposing effective structures of GANs. StackGAN \cite{c9} uses the stacked structure of multiple GANs to decompose the hard problem of generating high-resolution images into tractable subproblems. Subsequent studies \cite{c10,c11,c12,c13} have refined the architecture based on StackGAN \cite{c9}. AttnGAN \cite{c10} adopts cross-modal attention mechanisms for fine-grained generation. DM-GAN \cite{c11} leverages dynamic memory modules to supplement the generation procedure. Obj-GAN \cite{c12} and OP-GAN \cite{c13} use an additional input, pre-generated scene layout, to concentrate on creating objects. However, it is not easy to train multiple GANs at one time. Recently, DF-GAN \cite{c14} shows that a single pair of generator and discriminator can produce realistic images using deep fusion blocks.


\begin{figure}[t]
\begin{minipage}[b]{1.0\linewidth}
  \centerline{\includegraphics[width=\textwidth]{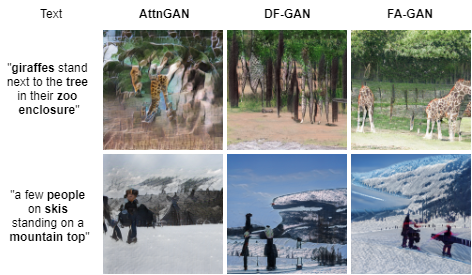}}
  \caption{Text-to-image synthesis examples of the proposed model FA-GAN and the baseline models AttnGAN, DF-GAN. The baseline models suffer from generating intact objects or details.}\medskip
\end{minipage}
\label{fig:fig1}
\vspace{-2.5em}
\end{figure}

\begin{figure*}[t]
\begin{minipage}[b]{1.0\linewidth}
  \centering
  \centerline{\includegraphics[width=\textwidth]{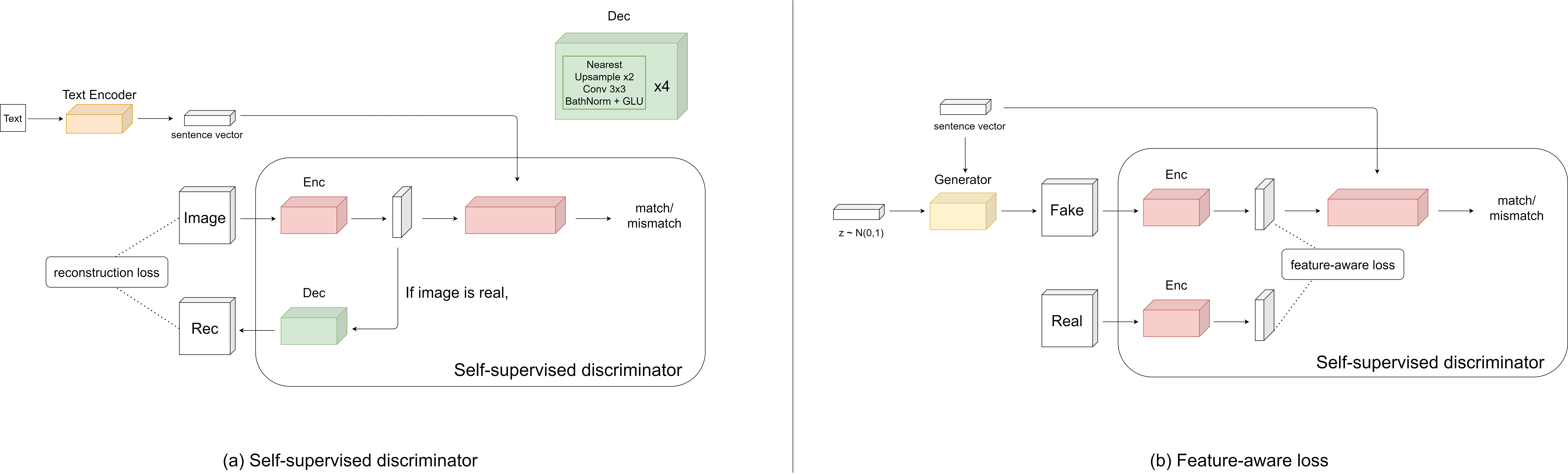}}
  \caption{The architecture of a proposed FA-GAN. (a) The self-supervised discriminator determines whether the input sentence and image match. The self-supervised discriminator has an auxiliary decoder, which is trained to reconstruct the real images. (b) The generator generates an image from the noise vector $z$ and sentence vector. The feature-aware loss maximizes the feature similarity between the fake image and the real image.}\medskip
\end{minipage}
\label{fig:fig2}
\vspace{-2.7em}
\end{figure*}


To encourage the text-image consistency, AttnGAN \cite{c10} suggests DAMSM loss and MirrorGAN \cite{mirrorgan} designs text-to-image-to-text cycle consistency loss. These methods compute the semantic similarity using a pre-trained network which slows down the training process. To improve this, DF-GAN \cite{c14} proposes a MA-GP which is a regularization method on the discriminator using the real images and does not require an extra network. On account of these efforts, the previous models can catch what to draw. However, they still struggle with generating intact objects or clear textures (Fig \ref{fig:fig1}). 

To circumvent this problem, we focus on the method to directly affect the generator by effectively exploiting the training data. To this end, we propose Feature-Aware Generative Adversarial Network (FA-GAN) to produce a more realistic image. Our method consists of two techniques, a self-supervised discriminator and a feature-aware loss. The self-supervised discriminator with an extra decoder extracts better feature representation by auto-encoding training. Feature-aware loss, which utilizes the representation, provides more direct supervision to the generator by indicating features that the generated image should have, which is a kind of regularization loss. The method can be easily applied to other existing models with small modification. Extensive experiments and ablation studies on the MS-COCO \cite{c16} dataset are conducted to demonstrate the superiority of FA-GAN. For the quantitative evaluation, we use Fr\'echet Instance Distribution (FID) \cite{c17}. Our proposed method advances the state-of-the-art FID from 28.92 to 24.58.

Overall, our contributions are as follows:
\begin {itemize}
\item We propose FA-GAN for the text-to-image synthesis, a method to give useful feedback to the generator to produce high-fidelity images.
\item Our proposed method is more effective than other similar regularization losses which use real data and generated data. 
\item FA-GAN outperforms the state-of-the-art models in terms of FID.
\end {itemize}

\section{Proposed Method}
\label{sec:method}

\subsection{Model Architecture}
\label{ssec:model}
We use the one stage GAN proposed in \cite{c14}. The overall architecture is illustrated in Fig \ref{fig:fig2}. It is composed of a pre-trained text encoder from \cite{c10}, a generator $G$, and a discriminator $D$. The pre-trained text encoder is a bi-directional Long Short-Term Memory that extracts a semantic vector from the text. The last hidden states are employed as the sentence vector $\bar{s}$. The generator $G$ generates an image $G(z,\bar{s})$ from the given sentence vector $\bar{s}$ and a noise vector $z$ sampled from Gaussian distribution $P_{z}$. With the addition of text condition, the role of the discriminator $D$ is different from the traditional discriminator which distinguishes whether the input image is real or fake. The discriminator $D$ has two inputs, a sentence vector $\bar{s}$ and an image $I$, and it determines whether the image and the text match or not. There are three cases to be considered while training $D$; (1) Match, when the input is a real image and matched sentence, (2) Mismatch, when the input is a real image and mismatched sentence, (3) Mismatch, when the input is a fake image regardless of the sentence. We use a hinge loss \cite{c18} to stabilize training GAN procedure and a MA-GP loss from \cite{c14}. The adversarial losses are defined as follows: 
\begin{equation}
    \begin{split}
    \mathcal{L}_{adv\_D} = & - \mathds{E}_{(x, \bar{s})\sim P_{data}}[\min(0,-1 + D(x, \bar{s}))]\\
        & -\frac{1}{2} \mathds{E}_{(\hat{x}, \bar{s}) \sim P_{mis}}[\min(0, -1 - D(\hat{x}, \bar{s}))]\\
        & -\frac{1}{2} \mathds{E}_{z \sim P_{z},\bar{s} \sim P_{data}}[\min(0, -1 - D(G(z,\bar{s}),\bar{s}))]\\
        & + \mathcal{L}_{MA-GP} \\
    \mathcal{L}_{adv\_G} = & - \mathds{E}_{z \sim P_{z},\bar{s} \sim P_{data}}[D(G(z, \bar{s}), \bar{s})]
    \end{split}
\end{equation}
where $(x,\bar{s})$ is the matched image and sentence from real distribution $P_{data}$, and $(\hat{x},\bar{s})$ is the mismatched image and sentence. 

\subsection{Self-supervised Discriminator}
\label{ssec:ssd}
Inspired by \cite{c15}, we design a self-supervised discriminator. We attach a single decoder $Dec$ to the intermediate layers of $D$ and denote the layers of the $D$ before the decoder as $Enc$. The $Dec$ has four convolution layers and the architecture is shown in the Fig \ref{fig:fig2}. The $Dec$ reconstructs the input image from the output of the $Enc$ and it is only optimized on the real samples during training $D$. We employ a perceptual loss \cite{c19} for the reconstruction loss:
\begin{equation}
    \mathcal{L}_{rec} = \mathds{E}_{(x,\bar{s})\sim P_{data}}[pl(Dec(Enc(x,\bar{s}))), x)] 
\end{equation}
where $pl$ is the perceptual loss function.
Such auto-encoding training makes $D$ to learn better feature representation from the inputs. To take an advantage of the representation, we utilize it for computing the feature-aware loss.

\subsection{Feature-Aware Loss}
\label{ssec:fa}

The self-supervised discriminator is prerequisite in feature-aware loss. It has an autoencoder internally since it contains an auxiliary decoder. Autoencoders are designed to encode the input into a meaningful representation. Thanks to the autoencoder structure in the self-supervised discriminator, we can regard the features of the real image $Enc(x,\bar{s})$ as the guidelines, allowing the generator to know which features the generated image should have. Consequently, we propose a \emph{feature-aware loss}, a regularization loss to synthesize an image that retains the features of a real image. Specifically, it is designed to enforce the $G$ to generate an image that maximizes the similarity of the feature representations between the generated image $G(z,\bar{s})$ and the corresponding real image $x$ with the same textual description $\bar{s}$. In this paper, we employ the $L^{1}$ distance to calculate the similarity and convert the maximization problem into a minimization problem. The proposed feature-aware loss is formulated as follows:

\begin{equation}
    \begin{split}
        \mathcal{L}_{fa} = \mathds{E}_{z \sim P_{z},(x,\bar{s})\sim P_{data}}[ |Enc(x,\bar{s}) - Enc(G(z,\bar{s}),\bar{s})| ]
    \end{split}
\end{equation}

In Sec \ref{ssec:losses}, we compare our method with perceptual loss. Perceptual loss  reconstructs the features obtained from the pre-trained VGG-16 \cite{c20} network. The main difference between our method and perceptual loss is which features to use for the regularization.

\subsection{Total Objective}
\label{ssec:subhead}

Finally, the total objective functions of our FA-GAN are defined as:
\begin{equation}
    \mathcal{L}_{D} = \mathcal{L}_{adv\_D} + \mathcal{L}_{rec}
\end{equation}
\begin{equation}
    \mathcal{L}_{G} = \mathcal{L}_{adv\_G} + \mathcal{L}_{fa}
\end{equation}

Our method can be easily applied to other models by adding the small number of parameters to the decoder $Dec$.

\setlength{\arrayrulewidth}{0.3mm}

\section{Experiments}
\label{sec:experiments}
We take extensive experiments to evaluate the performance of FA-GAN. We provide the quantitative and qualitative evaluation of FA-GAN against the state-of-the-arts on the MS-COCO dataset. In addition, we conduct ablation studies to investigate the effectiveness of the proposed method. Furthermore, we compare the feature-aware loss and other similar regularization losses. The results manifest the superiority of our proposed method.


\subsection{Evaluation Details}
\label{ssec:ed}
Inception Score (IS) \cite{c21} and Fr\'echet Inception Distance (FID) are widely used metrics for evaluating GANs. In contrast to IS, which evaluates only the distribution of generated images, the FID compares the distribution of generated images with the distribution of real images, which concludes that the FID is much more robust than IS. In other words, if the model generates the same image, the FID will be higher (the lower the FID, the better), but IS can not penalize this case. \cite{c12,c13,c14} found that IS is not an appropriate metric to evaluate the text-to-image synthesis models since some models tend to generate the same image when the text contains the same word, which is not good generative models but IS could be high (the higher the IS, the better). Thus, we use FID to evaluate our models. Following the prior works \cite{c11,c14}, we generate 30,000 images from randomly selected sentences in the test set and compute the FID score. 

\begin{table}[t]
    \centering
    \begin{tabular}{p{0.25\textwidth}|p{0.1\textwidth}}
    \hline
    Methods & FID \textdownarrow \\
    \hline
    AttnGAN & 35.49  \\
    DM-GAN & 32.64  \\
    DF-GAN & 28.92  \\
    FA-GAN (Ours) & \bf{24.58} \\
    \hline
    \end{tabular}
    \caption{Comparison with state-of-the-art text-to-image synthesis models. FA-GAN significantly outperforms the state-of-the-art. The best score is indicated in \bf{bold}.}
    \label{tab:sota}
\end{table}

\renewcommand{\arraystretch}{1.0}
\begin{table}[t]
    \centering
    \begin{tabular}{p{0.3\textwidth}|p{0.1\textwidth}}
    \hline
    Methods & FID \textdownarrow \\
    \hline
    baseline & 27.67  \\
    baseline + SSD & 37.32  \\
    baseline + FA$^-$ loss & 28.25  \\
    baseline + SSD + FA$^-$ loss (Ours) & \bf{24.58} \\
    \hline
    \end{tabular}
    \caption{Ablation study on our proposed method. SSD indicates the self-supervised discriminator and FA$^-$ indicates the feature-aware loss without the self-supervised discriminator.}
    \label{tab:abl}
    \vspace{-1.em}
\end{table}

\vspace*{-5pt}
\subsection{Quantitative Evaluation}
\label{ssec:quan}
We compare our model with state-of-the-art text-to-image synthesis models on MS-COCO dataset. As shown in the Table \ref{tab:sota}, the proposed FA-GAN significantly outperforms other methods. Compared with DF-GAN \cite{c14}, the FA-GAN improves the FID from 28.92 to 24.58. The results show that our proposed FA-GAN can generate images with better qualities.

\begin{figure}[t]
\centering
\includegraphics[width=8.5cm]{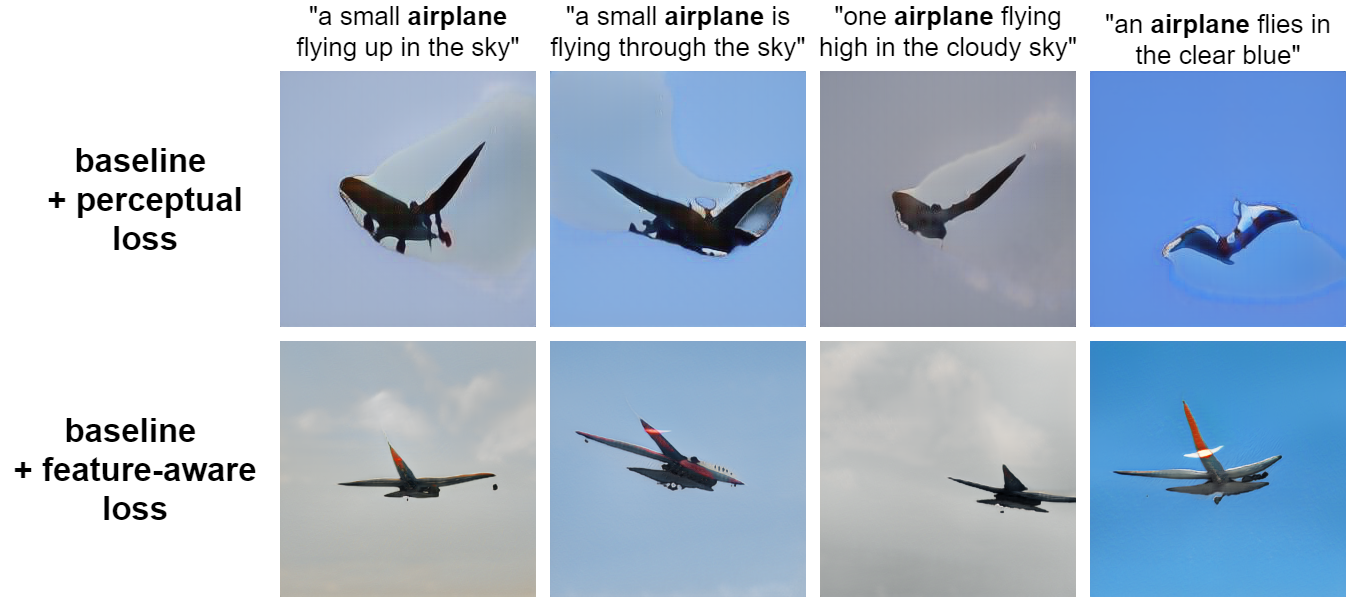}
\caption{Generated images from the similar sentences by \emph{baseline + perceptual loss} and \emph{baseline + feature-aware loss (Ours)}. Our method can generate more variants and clearer images.}
\vspace{-0.3em}
\label{fig:fig3}
\end{figure}

\begin{figure}[t]
\centering
\includegraphics[width=8.5cm]{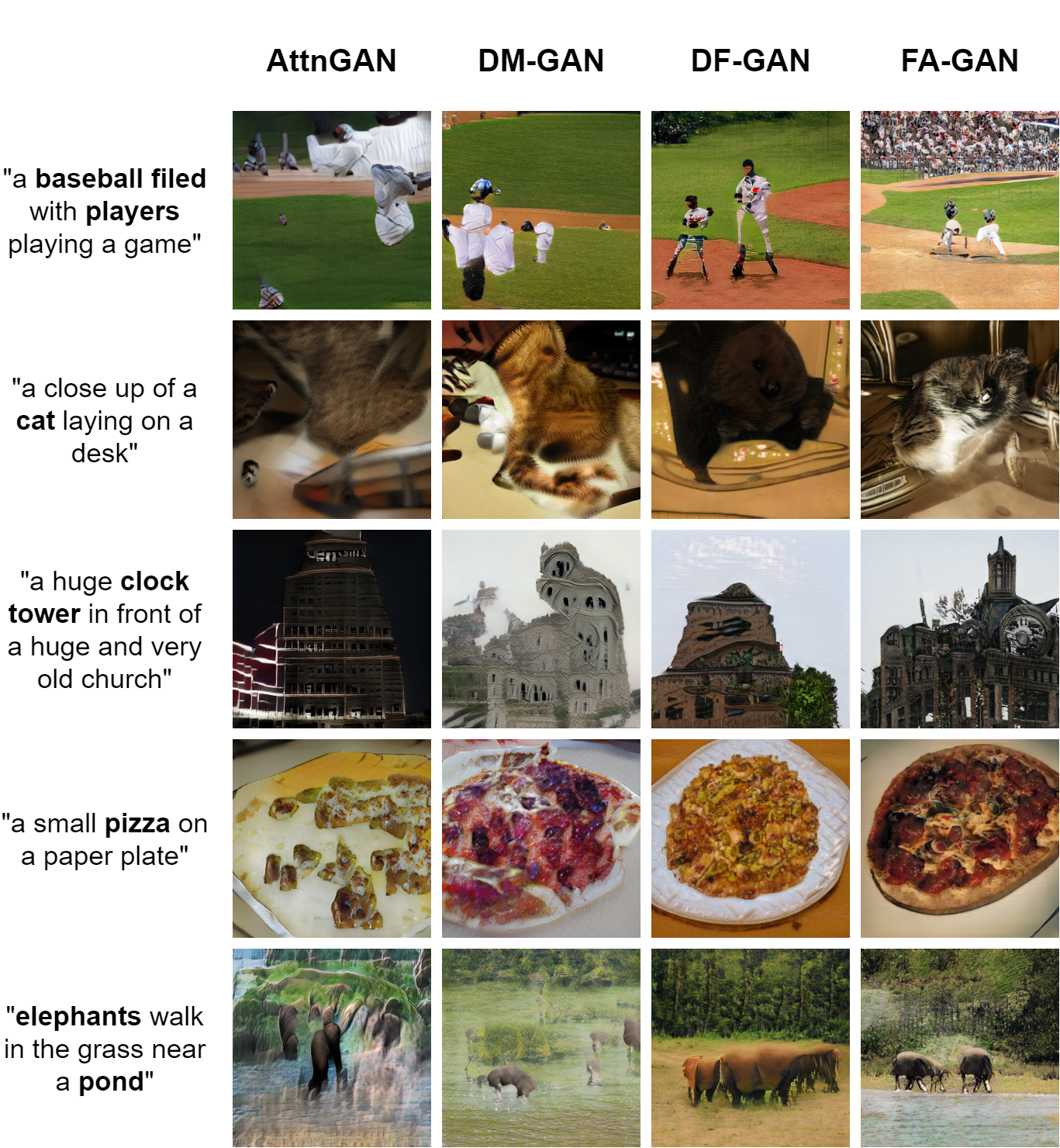}
\caption{Examples for the baseline models and FA-GAN on MS-COCO test set. FA-GAN generates more realistic images than the baseline models AttnGAN, DM-GAN, DF-GAN. }
\vspace{-1.em}
\label{fig:fig4}
\end{figure}

\vspace*{-5pt}
\subsection{Ablation Study}
\label{ssec:ab}
We perform ablation studies to verify the effectiveness of our proposed method. The components are the self-supervised discriminator (SSD) and the feature-aware loss (FA). We denote the variation that only uses feature-aware loss without SSD as FA$^-$. We compare four configurations, (1) FA-GAN w/o SSD and FA$^-$, (2) FA-GAN w/o FA$^-$, (3) FA-GAN w/o SSD, (4) FA-GAN. We set (1) as our baseline model which has a slightly lower FID score than DF-GAN \cite{c14}. The results are reported in Table \ref{tab:abl}. It shows that (2), (3) do not guarantee the improvement over the baseline. We found that the effect of (2) SSD without regularization accelerates the convergence speed of the discriminator, making the GAN training unstable and (3) has the side effects of regularization explained in Sec \ref{ssec:losses}. However, our proposed method (4) that integrates both SSD and FA$^-$ significantly improves the model's performance. We speculate that this is because regularizing meaningful representation can provide more useful signals to the generators.


\begin{table}[t]
    \centering
    \begin{tabular}{p{0.3\textwidth}|p{0.1\textwidth}}
    \hline
    Methods & FID \textdownarrow \\
    \hline
    baseline & 27.67 \\
    baseline + Perceptual loss & 37.84 \\
    baseline + FA$-$ loss& 28.25  \\
    baseline + FA loss (Ours) & \bf{24.58}  \\
    \hline
    \end{tabular}
    \caption{Experiments on the superiority of the feature-aware loss. Feature-aware loss is an effective regularization loss.}
    \label{tab:reg}
    \vspace{-1.0em}
\end{table}

\vspace*{-7pt}
\subsection{Comparison with other regularization methods}
\label{ssec:losses}
To verify the superiority of our proposed feature-aware loss, we compare the feature-aware loss with other similar regularization losses using feature representations. Perceptual loss uses the representation from the pre-trained VGG network for image classification, and FA$^-$ is from Sec \ref{ssec:ab}. The performances are reported in Table \ref{tab:reg}. Regularization loss is not always helpful for the GANs because it can hinder the diversity which causes the FID to increase. The side effects depend on the features which should contain meaningful information. We found that other regularization losses have the side effects, but our method has fewer side effects and more benefits. Fig \ref{fig:fig3} shows that our model can generate more variants and clear images. The results demonstrate that the feature-aware loss is beneficial regularization and our method can effectively exploit the training data. We conjecture that this is because the self-supervised discriminator extracts more meaningful high-level semantic features than others.

\vspace*{-10pt}
\subsection{Qualitative Results}
\label{ssec:subhead}
The Fig \ref{fig:fig4} shows the several examples of synthesized images by our proposed model FA-GAN and other baseline models. We observe that our model can retain the shape or generate clearer details compared to the baselines, which demonstrates the effectiveness of our proposed feature-aware loss.   

\section{Conclusion}
\label{sec:format}
In this paper, we propose FA-GAN for text-to-image synthesis which is a method to give the generator more useful signals. We design a self-supervised discriminator which is trained in an auto-encoding manner for extracting meaningful features. A feature-aware loss enables a model to generate images that have the same features as real images. Experimental results show the effectiveness of our method and our model can generate more realistic and clearer images. Moreover, the FA-GAN significantly advances the state-of-the-art FID from 28.92 to 24.58. The proposed method can be easily applied to other existing methods with only small modification.

\vspace{-11pt}

\section{ACKNOWLEDGEMENT}
This work was supported by Institute of Information \& communications Technology Planning \& Evaluation (IITP) grant funded by the Korea government(MSIT) (No.B0101-15-0266, Development of High Performance Visual BigData Discovery Platform for Large-Scale Realtime Data Analysis) and (No.2017-0-00897, Development of Object Detection and Recognition for Intelligent Vehicles)

\bibliographystyle{IEEEbib}
\bibliography{strings,refs}

\begin{thebibliography}{10}

\bibitem{c1}
Ian Goodfellow, Jean Pouget-Abadie, Mehdi Mirza, Bing Xu, David Warde-Farley,
  Sherjil Ozair, Aaron Courville, and Yoshua Bengio,
\newblock ``Generative adversarial networks,''
\newblock {\em Communications of the ACM}, vol. 63, no. 11, pp. 139--144, 2020.

\bibitem{c2}
Mehdi Mirza and Simon Osindero,
\newblock ``Conditional generative adversarial nets,''
\newblock {\em arXiv preprint arXiv:1411.1784}, 2014.

\bibitem{c3}
Phillip Isola, Jun-Yan Zhu, Tinghui Zhou, and Alexei~A Efros,
\newblock ``Image-to-image translation with conditional adversarial networks,''
\newblock in {\em Proceedings of the IEEE conference on computer vision and
  pattern recognition}, 2017, pp. 1125--1134.

\bibitem{c4}
Yunjey Choi, Youngjung Uh, Jaejun Yoo, and Jung-Woo Ha,
\newblock ``Stargan v2: Diverse image synthesis for multiple domains,''
\newblock in {\em Proceedings of the IEEE/CVF Conference on Computer Vision and
  Pattern Recognition}, 2020, pp. 8188--8197.

\bibitem{c6}
Albert Pumarola, Antonio Agudo, Aleix~M Martinez, Alberto Sanfeliu, and
  Francesc Moreno-Noguer,
\newblock ``Ganimation: Anatomically-aware facial animation from a single
  image,''
\newblock in {\em Proceedings of the European conference on computer vision
  (ECCV)}, 2018, pp. 818--833.

\bibitem{c7}
Jiguo Li, Xinfeng Zhang, Chuanmin Jia, Jizheng Xu, Li~Zhang, Yue Wang, Siwei
  Ma, and Wen Gao,
\newblock ``Direct speech-to-image translation,''
\newblock {\em IEEE Journal of Selected Topics in Signal Processing}, vol. 14,
  no. 3, pp. 517–529, Mar 2020.

\bibitem{c8}
Bowen Li, Xiaojuan Qi, Philip~HS Torr, and Thomas Lukasiewicz,
\newblock ``Lightweight generative adversarial networks for text-guided image
  manipulation,''
\newblock {\em arXiv preprint arXiv:2010.12136}, 2020.

\bibitem{c9}
Han Zhang, Tao Xu, Hongsheng Li, Shaoting Zhang, Xiaogang Wang, Xiaolei Huang,
  and Dimitris~N Metaxas,
\newblock ``Stackgan: Text to photo-realistic image synthesis with stacked
  generative adversarial networks,''
\newblock in {\em Proceedings of the IEEE international conference on computer
  vision}, 2017, pp. 5907--5915.

\bibitem{c10}
Tao Xu, Pengchuan Zhang, Qiuyuan Huang, Han Zhang, Zhe Gan, Xiaolei Huang, and
  Xiaodong He,
\newblock ``Attngan: Fine-grained text to image generation with attentional
  generative adversarial networks,''
\newblock in {\em Proceedings of the IEEE conference on computer vision and
  pattern recognition}, 2018, pp. 1316--1324.

\bibitem{c11}
Minfeng Zhu, Pingbo Pan, Wei Chen, and Yi~Yang,
\newblock ``Dm-gan: Dynamic memory generative adversarial networks for
  text-to-image synthesis,''
\newblock in {\em Proceedings of the IEEE Conference on Computer Vision and
  Pattern Recognition}, 2019, pp. 5802--5810.

\bibitem{c12}
Wenbo Li, Pengchuan Zhang, Lei Zhang, Qiuyuan Huang, Xiaodong He, Siwei Lyu,
  and Jianfeng Gao,
\newblock ``Object-driven text-to-image synthesis via adversarial training,''
\newblock in {\em Proceedings of the IEEE Conference on Computer Vision and
  Pattern Recognition}, 2019, pp. 12174--12182.

\bibitem{c13}
Tobias Hinz, Stefan Heinrich, and Stefan Wermter,
\newblock ``Semantic object accuracy for generative text-to-image synthesis,''
\newblock {\em arXiv preprint arXiv:1910.13321}, 2019.

\bibitem{c14}
Ming Tao, Hao Tang, Songsong Wu, Nicu Sebe, Fei Wu, and Xiao-Yuan Jing,
\newblock ``Df-gan: Deep fusion generative adversarial networks for
  text-to-image synthesis,''
\newblock {\em arXiv preprint arXiv:2008.05865}, 2020.

\bibitem{mirrorgan}
Tingting Qiao, Jing Zhang, Duanqing Xu, and Dacheng Tao,
\newblock ``Mirrorgan: Learning text-to-image generation by redescription,''
\newblock in {\em Proceedings of the IEEE/CVF Conference on Computer Vision and
  Pattern Recognition}, 2019, pp. 1505--1514.

\bibitem{c16}
Tsung-Yi Lin, Michael Maire, Serge Belongie, James Hays, Pietro Perona, Deva
  Ramanan, Piotr Doll{\'a}r, and C~Lawrence Zitnick,
\newblock ``Microsoft coco: Common objects in context,''
\newblock in {\em European conference on computer vision}. Springer, 2014, pp.
  740--755.

\bibitem{c17}
Martin Heusel, Hubert Ramsauer, Thomas Unterthiner, Bernhard Nessler, and Sepp
  Hochreiter,
\newblock ``Gans trained by a two time-scale update rule converge to a local
  nash equilibrium,''
\newblock {\em Advances in Neural Information Processing Systems}, vol. 30, pp.
  6626--6637, 2017.

\bibitem{c18}
Jae~Hyun Lim and Jong~Chul Ye,
\newblock ``Geometric gan,''
\newblock {\em arXiv preprint arXiv:1705.02894}, 2017.

\bibitem{c15}
Bingchen Liu, Yizhe Zhu, Kunpeng Song, and Ahmed Elgammal,
\newblock ``Towards faster and stabilized gan training for high-fidelity
  few-shot image synthesis,''
\newblock {\em arXiv e-prints}, pp. arXiv--2101, 2021.

\bibitem{c19}
Justin Johnson, Alexandre Alahi, and Li~Fei-Fei,
\newblock ``Perceptual losses for real-time style transfer and
  super-resolution,''
\newblock in {\em European conference on computer vision}. Springer, 2016, pp.
  694--711.

\bibitem{c20}
Karen Simonyan and Andrew Zisserman,
\newblock ``Very deep convolutional networks for large-scale image
  recognition,''
\newblock {\em arXiv preprint arXiv:1409.1556}, 2014.

\bibitem{c21}
Tim Salimans, Ian Goodfellow, Wojciech Zaremba, Vicki Cheung, Alec Radford, and
  Xi~Chen,
\newblock ``Improved techniques for training gans,''
\newblock {\em arXiv preprint arXiv:1606.03498}, 2016.

\end{thebibliography}
\end{document}